\documentclass[letterpaper, 10 pt, conference]{ieeeconf}  

\IEEEoverridecommandlockouts                              

\usepackage{epsfig} 
\usepackage{amsmath} 
\usepackage{amssymb}  
\usepackage{url}
\usepackage{booktabs} 
\usepackage{cite}
\usepackage{subcaption}
\usepackage{caption}
\usepackage{graphicx}
\usepackage{xcolor}         
\usepackage{comment}
\usepackage{calc}
\newcommand{\levelname}[1]{\hspace{10em}\makebox[\widthof{LEVEL 5 (DEFORMABILITY)}][l]{#1}}
\usepackage[ruled,linesnumbered]{algorithm2e}
\usepackage{multirow}
\usepackage{balance}
\usepackage{hyperref}
\usepackage{setspace}

\makeatletter
\newcommand{\removelatexerror}{\let\@latex@error\@gobble}
\makeatother

\newcommand{\cut}[1]{}

\newcommand{\para}[1]{{\noindent\textbf{#1}}}
\newcommand{\parai}[1]{{\noindent\textit{#1}}}

\newcommand{\gt}[1]{\textcolor{black}{#1}}
\newcommand{\nonl}{\renewcommand{\nl}{\let\nl\oldnl}} 

\title{\LARGE \bf MOSAIC: Learning Unified Multi-Sensory Object Property Representations for Robot Learning via Interactive Perception}

\author{
    \authorblockN{Gyan Tatiya$^{1*}$\thanks{*Work done during internship at Bosch Center for Artificial Intelligence.} \quad Jonathan Francis$^{2}$  \quad Ho-Hsiang Wu$^{2}$ \quad Yonatan Bisk$^{3}$ \quad Jivko Sinapov$^{1}$}
    \thanks{$^{1}$Department of Computer Science, Tufts University, Email: {\tt\footnotesize \{Gyan.Tatiya, Jivko.Sinapov\}@tufts.edu}.
    $^{2}$Bosch Center for AI, Email:
    {\tt\footnotesize \{Jon.Francis, Ho-Hsiang.Wu\}@us.bosch.com}.
    $^{3}$Carnegie Mellon University, Email:
    {\tt\footnotesize ybisk@andrew.cmu.edu.}}
    \vspace{-1cm}
}

\begin{document}

\maketitle
\thispagestyle{empty}
\pagestyle{empty}

\begin{abstract}

A holistic understanding of object properties across diverse sensory modalities (e.g., visual, audio, and haptic) is essential for tasks ranging from object categorization to complex manipulation.
Drawing inspiration from cognitive science studies that emphasize the significance of multi-sensory integration in human perception, we introduce MOSAIC (Multimodal Object property learning with Self-Attention and Interactive Comprehension), a novel framework designed to facilitate the learning of unified multi-sensory object property representations.
While it is undeniable that visual information plays a prominent role, we acknowledge that many fundamental object properties extend beyond the visual domain to encompass attributes like texture, mass distribution, or sounds, which significantly influence how we interact with objects.
In MOSAIC, we leverage this profound insight by distilling knowledge from multimodal foundation models and aligning these representations not only across vision but also haptic and auditory sensory modalities.
Through extensive experiments on a dataset where a humanoid robot interacts with 100 objects across 10 exploratory behaviors, we demonstrate the versatility of MOSAIC in two task families: object categorization and object-fetching tasks.
Our results underscore the efficacy of MOSAIC's unified representations, showing competitive performance in category recognition through a simple linear probe setup and excelling in the fetch object task under zero-shot transfer conditions.
This work pioneers the application of sensory grounding in foundation models for robotics, promising a significant leap in multi-sensory perception capabilities for autonomous systems.
We have released the code, datasets, and additional results: \footnotesize\texttt{\href{https://github.com/gtatiya/MOSAIC}{https://github.com/gtatiya/MOSAIC}}.

\end{abstract}

\section{Introduction}

Humans first acquire knowledge about object properties through physical interaction---a process that involves the integration of multiple sensory inputs, including visual, auditory, and tactile cues \cite{thesen2004neuroimaging, alais2010multisensory, bulkin2006seeing, kim2023robotic, zhang2023multimodal, li2020review, tatiya2023transferring}.
For instance, we rely on vision to discern an object's color, sense of touch when we lift an object to gauge its weight, and hearing when we shake a container to determine if it is full or empty.
The fusion of such multi-sensory information is pivotal in shaping our perception and guiding our decision-making processes concerning objects \cite{bizley2016multisensory, parise2012correlation, francis2022core, hu2023toward, chen2023augmented, cai2021visual}.
Similarly, robots can effectively engage with objects by simultaneously perceiving and processing multi-sensory signals, to tackle tasks such as object categorization \cite{sinapov_grounding_2014, tatiya2019deep}, material recognition \cite{xiong2022deeply}, and even complex actions like packing and pouring \cite{li_see_2022}.

Within vision and text, large-scale Vision-Language Models (VLMs) have demonstrated their ability to provide state-of-the-art representations for both visual and textual modalities, making them exceptionally valuable for a wide range of AI applications \cite{radford_learning_2021, zhang2022contrastive, alayrac2022flamingo}.
One such model, Contrastive Language-Image Pre-training (CLIP) \cite{radford_learning_2021}, is trained from scratch on an extensive dataset comprising 400 million (image, text) pairs.
CLIP's representations can seamlessly transfer to many downstream tasks without fine-tuning.
While prior research has primarily focused on integrating audio modalities into CLIP's embedding space \cite{wu_wav2clip_2022}, including a robot's haptic data into this versatile space has yet to be explored.
We address this gap by distilling the domain-general language grounding within CLIP and infusing it into a robot's sensory data from object interactions.
This method effectively mitigates the often prohibitive costs of collecting interactive data by robots through extensive object exploration.
The primary objective of this study is to expose VLMs to object property representations derived from robot interactions, highlighting how these representations can significantly improve the performance on interactive tasks by enhancing the robot's multimodal perceptual capabilities.
This enhancement arises from interactive object exploration to understand the fundamentals of object properties, a perspective disembodied representations often lack.

We introduce our method and framework for learning \textbf{M}ultimodal \textbf{O}bject properties with \textbf{S}elf-\textbf{A}ttention and \textbf{I}nteractive \textbf{C}omprehension  (MOSAIC)---an approach for acquiring versatile representations that are adaptable to various interactive perception tasks within robotics.
MOSAIC is designed to extract unified multi-sensory object property representations, enabling understanding of object properties by leveraging diverse sensory modalities.
This approach rests on the premise that natural language provides a versatile embedding space 
whose knowledge we can distill 
and align 
to different sensory modalities.
We evaluate 
our approach on 
a publicly available dataset where a humanoid robot explored 100 objects, using 10 exploratory behaviors while recording sensory data, including vision, audio, and haptic.
We evaluate on both object category recognition and the fetch object task, finding MOSAIC to be robust and adaptible.
MOSAIC's performance in the object category recognition is notably competitive compared to state-of-the-art methods, showing the effectiveness of unified representations even within a straightforward linear probe setup.
Furthermore, MOSAIC demonstrates exceptional capabilities in executing natural language instructions in the fetch object task under a zero-shot condition.
In summary, MOSAIC offers a versatile framework for multimodal object property learning, bridging the gap between different sensory inputs and facilitating a wide range of downstream robot tasks.

\begin{figure*}[ht]
\centering
\includegraphics[width=\linewidth]{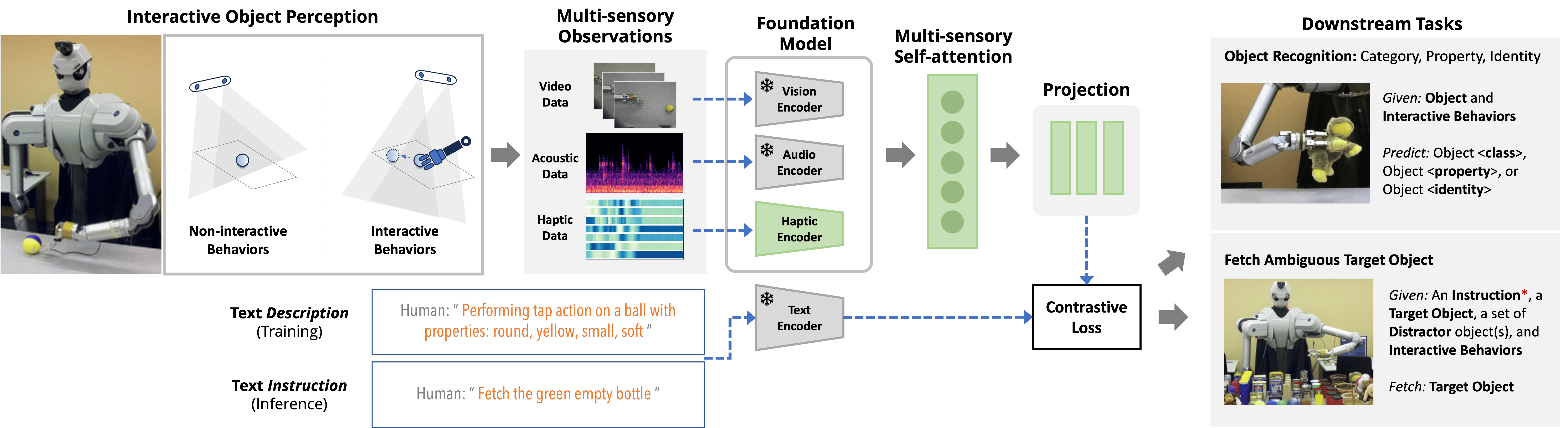}
\vspace{-0.6cm}
\caption{\small \textbf{Overview of the MOSAIC Framework}: Initially, the robot collects sensory data through object exploration, which is then used to train models for distilling unified multimodal representations guided by a pre-trained text encoder. These acquired representations are subsequently applied to a variety of downstream tasks.}
\label{fig:System_overview}
\vspace{-0.6cm}
\end{figure*}

\section{Related Work}
\vspace{-0.2cm}
\para{Multi-sensory Learning in Cognitive Science.}
Humans acquire knowledge about object properties through physical interactions, integrating multiple sensory signals \cite{borghi2002role, lacey2014visuo, calvert2004multisensory}.
Multi-sensory integration and attention processes occur at various stages in the human brain, crucially influencing our perception of objects and task performance \cite{koelewijn2010attention}.
Moreover, human perception involves the dynamic interplay between sensory inputs and existing cognitive knowledge rather than processing sensory inputs in isolation \cite{talsma2015predictive}.
Our research extends these principles to robotics, extracting knowledge from pre-trained text encoders to align representations across diverse sensory modalities---
mirroring how humans fuse sensory information with their established knowledge to perceive their environment holistically.

\para{Interactive Perception.}
Robotics research has showcased the remarkable capabilities of robots in interacting with objects and leveraging sensory signals for an array of tasks, encompassing object categorization \cite{sinapov_grounding_2014, tatiya2019deep, tatiya2019sensorimotor, tatiya2020haptic, tatiya2020framework}, material recognition \cite{xiong2022deeply}, and intricate manipulation actions like packing and pouring \cite{li_see_2022}.
Most successful prior work relies on 
handcrafted auditory, haptic, and visual features \cite{sinapov_grounding_2014}, or 
specialized 
architectures for processing raw multi-sensory data to predict object categories \cite{tatiya2019deep}.
Recently, 
Li {\it et al.} \cite{li_see_2022}, introduced a self-attention model 
to fuse information from visual, auditory, and tactile sensors, significantly enhancing the robot's capability to tackle complex manipulation tasks.
Our research introduces a versatile framework for learning unified multi-sensory representations \emph{from raw sensory data} acquired during robot-object interactions, offering adaptability across diverse downstream tasks.
The generality of our network architecture has been 
is demonstrated across various applications with strong performance, and the inclusion of self-attention mechanisms further bolsters its performance.

\para{Unified Multi-Sensory Representations with CLIP.} Recent advances 
have revealed the potential of contrastive objectives to yield generalized representations for both text and images \cite{radford_learning_2021, zhang2022contrastive}.
Contrastive Language-Image Pre-training (CLIP) \cite{radford_learning_2021} has delivered state-of-the-art representations that excel in diverse tasks, including zero-shot image classification, image retrieval via text, and guiding generative models \cite{gal2022stylegan}.
While 
CLIP's knowledge has been distilled for 
audio \cite{wu_wav2clip_2022}, 
our MOSAIC approach is the first to 
ground sensory data obtained through robotic object exploration. 
MOSAIC accomplishes this by distilling knowledge from the extensive pre-trained CLIP text model.
To test our learned unified representations, we rely on a dataset where a robot engages with 100 objects, executing 10 exploratory behaviors while recording multiple sensory signals.
The robot tackles two tasks reliant on perceiving object properties: object categorization and the fetch object task.
The results highlight the efficiency of our unified representations, clearly demonstrated in competitive performance in category recognition only by using a simple linear probe setup and in fetch object task using a zero-shot transfer approach.

\vspace{-0.2cm}
\section{Learning Methodology}
\label{sec:approach}
\vspace{-0.1cm}
\para{Notation and Problem Formulation.} Let a robot perform a set of exploratory behaviors $\mathcal{B}$ (e.g., {\it grasp}, {\it pick}) on a set of household objects $\mathcal{O}$ (e.g., {\it bottle}, {\it cup}), while recording a set of sensory modalities, $m = \{\mathrm{x}^{v}, \mathrm{x}^{a}, \mathrm{x}^{h}\}$, which correspond to {\it vision}, {\it audio}, {\it haptics}, respectively.
The robot performs each behavior $n$ times on each object.
During the $i^{th}$ exploratory trial, the robot collects sensory data $m_i$ containing:
\begin{equation}
    \mathrm{x}_{i}^{v} \in \mathbb{R}^{w\times h \times 3 \times t_{i}^{v}},
    \mathrm{x}_{i}^{a} \in \mathbb{R}^{f\times t_{i}^{a}},
    \mathrm{x}_{i}^{h} \in \mathbb{R}^{d \times t_{i}^{h}}
\label{eq:robot_observe}
\end{equation}
\noindent where $w$ and $h$ are the width and height of each image, $f$ is the number of frequency bins in the sound spectrogram, $d$ is the number of robot joint-torque sensors, and $t_{i}^{v}$, $t_{i}^{a}$, and $t_{i}^{h}$ are the number of time frames (e.g., number of images) produced during interaction for vision, audio, and haptics, respectively.
Additionally, the robot has access to textual descriptions of each interaction, $\mathrm{x}_{i}^{s}$, provided by human experts, complementing the sensory data.

Our primary objective is to learn a unified multimodal representation derived from the robot's observations across all modalities during an exploratory trial.
To be more precise, we aim to learn the function $F_{m \rightarrow \mathcal{Z}}:x_{i}^{v}, x_{i}^{a}, x_{i}^{h} \rightarrow z_i$, where $z_i \in \mathbb{R}^{D_{\mathcal{Z}}}$ represents the unified multimodal embedding of dimension $D_{\mathcal{Z}}$.
This unified representation is intended to encompass diverse object properties encountered during interactions, making it applicable to various downstream tasks that require understanding these object properties.
By achieving this unified representation, the robot can rapidly adapt to different tasks by learning linear models or performing zero-shot transfers, thereby circumventing the need to train complex models dedicated to individual tasks.

\para{Learning Unified Multimodal Object Properties.}
Our approach, MOSAIC (\textbf{M}ultimodal \textbf{O}bject property learning with \textbf{S}elf-\textbf{A}ttention and \textbf{I}nteractive \textbf{C}omprehension), involves a two-stage process, illustrated in Fig. \ref{fig:System_overview}.
Initially, we aim to distill unified object property representations from diverse sensory modalities, guided by text embeddings from a pre-trained text encoder.
Subsequently, we leverage these unified representations to solve downstream tasks that require understanding object properties.
In the following sections, we introduce various modules integrated within our framework.

\subsubsection{Encoders and Feature Extraction}

For the \texttt{Vision Encoder}, we use the CLIP's Vision Transformer (ViT-B/32) \cite{radford_learning_2021}, which is jointly trained with a text encoder to maximize the similarity of \{image, text\} pairs using a contrastive loss. For each interaction's video, the image encoder extracts image embeddings, and these embeddings are then aggregated using adaptive average pooling to generate a feature vector of size $D_{\mathcal{Z}}$.
For the \texttt{Audio Encoder}, we leverage the Wav2CLIP model \cite{wu_wav2clip_2022}
, which is trained to project audio data into the shared vision-language embedding space of CLIP; this approach enables the extraction of audio embeddings of size $D_{\mathcal{Z}}$.
For the \texttt{Haptics Encoder}, we use a ResNet-18 \cite{he2016deep} model, pre-trained on the ImageNet dataset, as the foundation. The input channels of the first convolutional layer are modified to one channel, and the output of the last fully-connected layer is adapted to match the desired embedding size of $D_{\mathcal{Z}}$; a sample haptic image is shown in Fig. \ref{fig:System_overview}.
\texttt{Text Encoder}: For each exploratory trial, a corresponding natural language description is available. Leveraging CLIP's text encoder (ViT-B/32) \cite{radford_learning_2021}, we extract embeddings of size $D_{\mathcal{Z}}$ from these text descriptions.

\SetKwData{visionencoder}{vision\_encoder}%
\SetKwData{audioencoder}{audio\_encoder}%
\SetKwData{hapticencoder}{haptic\_encoder}%
\SetKwData{textencoder}{text\_encoder}%
\SetKwData{concatenation}{concatenation}%
\SetKwData{multiheadattention}{multihead\_attention}%
\SetKwData{MLPencoder}{MLP\_encoder}%
\SetKwData{exp}{exp}%
\SetKwData{range}{range}%
\SetKwData{crossentropyloss}{cross\_entropy\_loss}%
\begin{figure}[t]
\removelatexerror
\begin{algorithm}[H]
\footnotesize
  \caption{{\tt\footnotesize Training MOSAIC Framework}}
  \label{alg:training_MOSAIC_framework}
  {\nonl {\boldmath $V, A, H, S$}: Minibatch of aligned data (vision, audio, haptic, text)} \\
  {\nonl {\boldmath $n$}: Size of minibatch} \\
  {\nonl {\boldmath $MOSAIC_\theta$}: Learnable parameters of MOSAIC framework} \\

  {\nonl \normalfont{// Extract feature vector for each modality}} \\
  $V_f$ = \visionencoder{$V$} \hfill \normalfont{// Vision Transformer} \\
  $A_f$ = \audioencoder{$A$} \hfill \normalfont{// Wav2CLIP model} \\
  $H_f$ = \hapticencoder{$H$} \hfill \normalfont{// ResNet18 model} \\
  $S_f$ = \textencoder{$S$} \hfill \normalfont{// Text Transformer} \\

  {\nonl \normalfont{// Compute unified representation}} \\
  $U_f$ = \concatenation{$V_f, A_f, H_f$} \\
  $U_f$ = \multiheadattention{$U_f$} \\
  $U_f$ = \MLPencoder{$U_f$} \hfill \normalfont{// MLP model} \\

  {\nonl \normalfont{// Scaled pairwise cosine similarities}} \\
  $logits$ = $U_f \cdot S_f^\top $ \\

  {\nonl \normalfont{// Symmetric loss function}} \\
  $labels$ = \range{$n$} \hfill \normalfont{// returns 1, 2, ..., $n$} \\
  $loss_u$ = \crossentropyloss{$labels, logits$} \\
  $loss_s$ = \crossentropyloss{$labels, logits^\top$} \\
  $loss$ = $(loss_u + loss_s)/2$ \\

  Update $MOSAIC_\theta$ to minimize $loss$
\end{algorithm}
\vspace{-0.7cm}
\end{figure}

\subsubsection{Multimodal Fusion}

We employ a self-attention mechanism to integrate the feature sets from the three modalities. Beginning with the concatenation of feature vectors from each modality, we apply a two-step process:
first, conventional multi-head self-attention \cite{vaswani2017attention} is applied to the concatenated features; subsequently, the resulting output is directed through a Multi-Layer Perceptron (MLP) to yield the unified multi-sensory feature of size $D_{\mathcal{Z}}$.

\subsubsection{Training}

During training, we maintain the vision, audio, and text encoders in their frozen states since they were already tuned to project into a shared embedding space.
We train the haptic, self-attention, and MLP networks.
Our primary aim is to create unified multimodal representations within the same embedding space as CLIP's text embeddings \cite{radford_learning_2021}.
To accomplish this, we employ a distillation method guided by CLIP's text embeddings.
We follow the approach outlined in the original CLIP paper, using a contrastive loss mechanism.
This involves employing positive examples from different modalities within the same data sample while considering negative examples from the remaining batch.
The fundamental implementation of this training process is shown in Algorithm \ref{alg:training_MOSAIC_framework}.
This strategy is predicated on the concept that natural language offers a versatile grounding basis \cite{bisk2020experience}, facilitating the creation of generalized representations with effective transferability across diverse downstream tasks.

\vspace{-0.2cm}
\section{Experimental Design}
\begin{figure}
\centering
\includegraphics[width=0.95\linewidth]{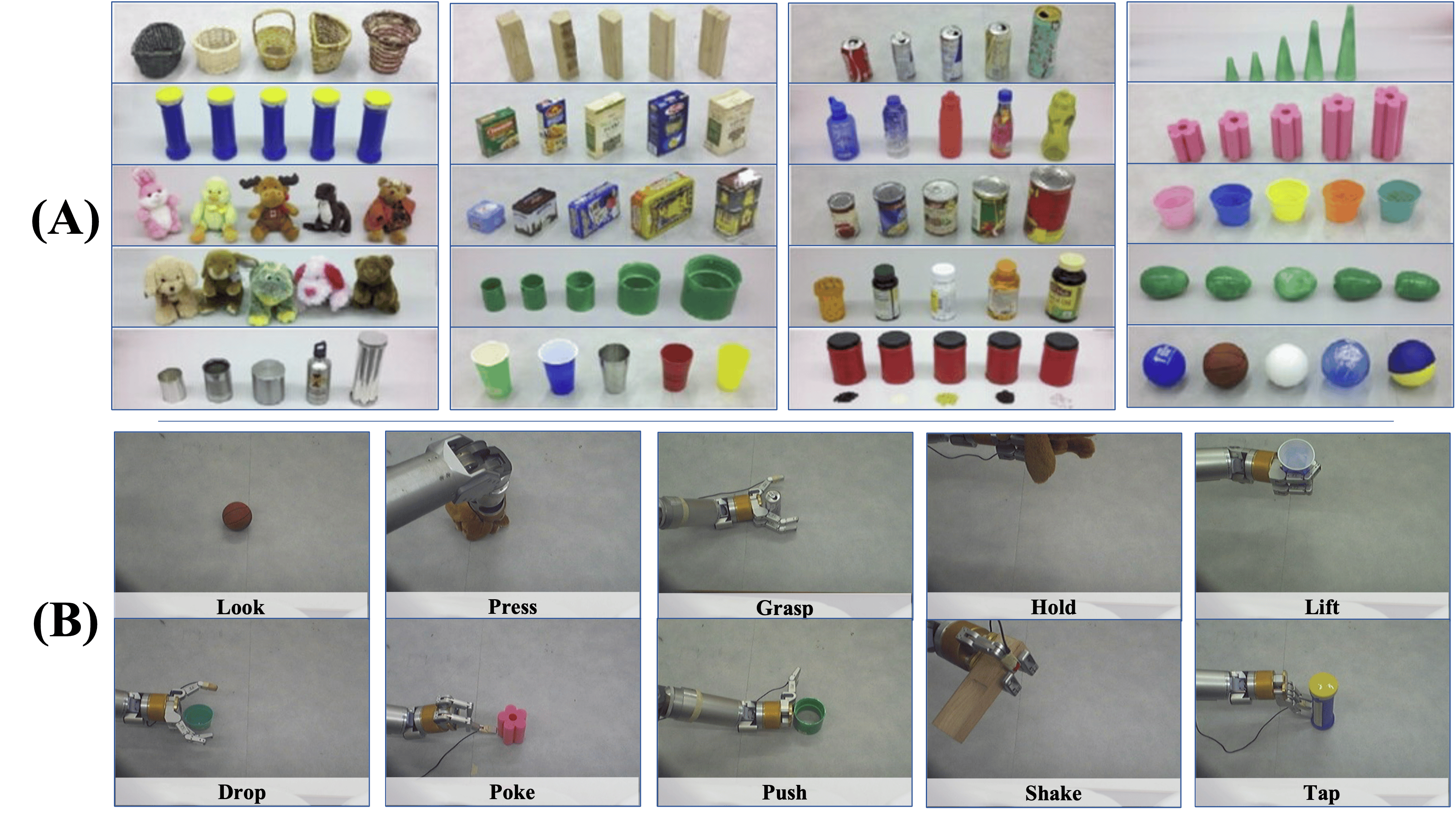}
\vspace{-0.2cm}
\caption{\small (A) 100 objects, grouped in 20 object categories. (B) The interactive behaviors that the robot performed on the objects.
}
\label{fig:Objects_Behaviors}
\vspace{-0.7cm}
\end{figure}

\vspace{-0.1cm}
\para{Sensory Dataset.} We used the publicly accessible dataset collected by Sinapov {\it et al.} \cite{sinapov_grounding_2014}.
In this experiment, a humanoid robot (depicted in Fig. \ref{fig:System_overview}) explored 100 household objects from 20 different categories (shown in Fig. \ref{fig:Objects_Behaviors}A), using 10 exploratory behaviors. 
These behaviors included {\it Look}, {\it Press}, {\it Grasp}, {\it Hold}, {\it Lift}, {\it Drop}, {\it Poke}, {\it Push}, {\it Shake}, and {\it Tap} (shown in Fig. \ref{fig:Objects_Behaviors}B).
{\it Look} is a non-interactive behavior, only capturing visual data.
For every interactive behavior, the robot collected sensory data including visual, audio, and haptic, acquired through three sensors:
(1) A Logitech webcam capturing 320 x 240 RGB images at 10 frames per second;
(2) An Audio-Technica U853AW cardioid microphone capturing audio sampled at 44.1 KHz;
(3) Joint-torque sensors capturing torques from all 7 joints at 500 Hz.
The robot repeated each behavior 5 times for each of the 100 objects, resulting in a total of 5,000 interactions (10 behaviors x 5 trials x 100 objects).

\para{Text Dataset.} The objects in our dataset were annotated with properties, shown in Table \ref{tab:property_categories}, each with corresponding values.
While not all properties were applicable to every object (e.g., the {\it baseball} object lacked a weight property), we leveraged these properties to generate text descriptions for each interaction.
To ensure diversity, we randomly selected a subset of properties for each object and used them in the descriptions.
For each object's text description, we ensured that it included at least one property, and the maximum number of properties included was determined by the number of properties with values for that object.
Moreover, we included the behavior's name being executed (e.g., {\it tap}), the object's category (e.g., {\it ball}), and the category of different object properties (e.g., {\it material}), all chosen randomly.
Further variety was introduced by selecting synonyms for words within the description from a curated set of synonyms corresponding to the dataset's labels.
We generated 100 unique text descriptions using this random selection process for each combination of object and behavior.
For instance, an example text description might read: {\it ``Performing tap action on a ball with properties: round, yellow, small, soft, toy''}.


\para{Data Pre-processing.} To ensure synchronization and consistency across all sensory modalities for each behavior $b \in \mathcal{B}$, we calculated the behavior's duration by dividing the average number of images recorded during behavior $b$ by 10 (camera's frame rate).
With the duration of each behavior now fixed, we compute the average number of time frames for each modality by multiplying this duration and the frame rate specific to that modality.
These calculated averages were used for interpolation, ensuring uniform time frames for each modality during the interaction recording.
For images and text, we employed the pre-processing provided by CLIP \cite{radford_learning_2021}.
Audio data was transformed from raw waveforms (1D) to spectrograms (2D) using the audio preprocessor from Wav2CLIP \cite{wu_wav2clip_2022}.
For haptic signals, we applied dimensionality reduction by interpolating the original 500Hz sampling rate down to 50Hz, drawing inspiration from a similar technique used in a prior study \cite{tatiya2019deep} conducted with the same dataset we used in our experiments.

\para{Model Implementation.} We standardized the size of the embeddings at $D_{\mathcal{Z}}$ = 512. Our framework was implemented in PyTorch \cite{paszke2019pytorch}, which includes the multi-sensory self-attention model and MLP encoder. 

\para{Validation Procedure.} 
Each of the 20 object categories consists of 5 unique objects.
To train our framework, we selected 4 objects from each category for the training set while reserving one object for testing, resulting in a training set with 80 objects and a testing set with 20 objects.
We employed a 5-fold object-based cross-validation strategy to ensure that each object appeared four times in the training set and once in the test set.
Given that the robot interacted with each object 5 times, our training set contained 400 examples (80 objects $\times$ 5 trials); the test set comprised 100 examples (20 objects $\times$ 5 trials) for each exploratory behavior.
\gt{During training, we randomly select a text description for the given object and behavior from the corresponding pool of 100 text descriptions, ensuring variability in the training process.}

\SetKwData{textencoder}{text\_encoder}%
\SetKwData{performbehavior}{perform\_behavior}%
\SetKwData{getunifiedrepr}{get\_unified\_repr}%
\SetKwData{cosinesimilarity}{cosine\_similarity}%
\begin{algorithm}[t]
\footnotesize
  \caption{{\tt\footnotesize Fetch\_object($c,\ O,\ B,\ \theta$)}}
  \label{alg:fetch_object}
  {\nonl {\boldmath $MOSAIC_\theta$}: Learned parameters in Algorithm \ref{alg:training_MOSAIC_framework}} \\

  $t_c$ = \textencoder{c}: \hfill \normalfont{// Command to fetch target} \\

  \For{$o \in O$: \normalfont{Set of objects (target and distractor(s))}}{
    $similarity$ = $0$ \\
    \For{$b \in B$: \normalfont{Set of Behaviors}}{
    $sensory\_data$ = \performbehavior{o, b} \\
    $u_b$ = \getunifiedrepr{$sensory\_data, MOSAIC_\theta$} \\
    $similarity$ += \cosinesimilarity{$t_c, u_b$}
    }
    Save $similarity$ for $o$
  }
  \Return{\normalfont{Target Object} $o$ \normalfont{with highest $similarity$}}
\end{algorithm}
\begin{table}[t]
\vspace{-0.3cm}
\caption{\small Property categories and associated descriptive words.}
\vspace{-0.15cm}
\label{tab:property_categories}
\scriptsize
\centering
\resizebox{0.9\columnwidth}{!}{
\begin{tabular}{@{}l@{\hspace{4pt}}p{\linewidth}@{}}
\toprule
{\it \textbf{Properties}} & {\it \textbf{Values}} \\
\midrule
Color & brown, blue, pink, red, white, orange, yellow, green, purple, multicolored \\
Deform. & deformable, rigid, brittle \\
Hardness & soft, squishy, hard \\
Material &  plastic, wicker, aluminum, foam, metal, rubber, paper, styrofoam, wood \\
State & closed, full, empty, open \\
Reflection & shiny, dull \\
Shape & cylindrical, wide, rectangular, block, box, cone, round \\
Size & small, short, big, large, tall \\
Transp. & transparent, opaque, translucent, see-through \\
Usage & container, toy \\
Weight & light, heavy \\
\bottomrule
\end{tabular}
}
\vspace{-0.6cm}
\end{table}

\para{Evaluation Tasks.} After training our framework, we extracted the unified representations by freezing learned weights for all downstream tasks. We evaluated the acquired representations through two distinct tasks.
The following subsections elaborate on these tasks, outlining our approach to tackling them with unified representations and discussing our performance metrics.
Additionally, we discuss the baseline methods we employed for comparison with our method.

\subsubsection{Object Category Recognition}

In this task, the robot interacts with a given object to identify its category from a set of 20 categories.
We use a standard multi-class linear classifier for supervised classification.
Specifically, we use a Multi-Layer Perceptron (MLP) architecture that takes the unified representation as input, passes it through a hidden layer and a ReLU activation function, and produces 20 logits 
for 20 categories.
We train this classifier using the cross-entropy loss function for 50 epochs, using the Adam optimizer \cite{kingma_adam_2015} with a learning rate of $10^{-4}$.
The trained classifier is then used to recognize the category of test objects, and we compute accuracy as a performance metric, defined as $A = \frac{\text{correct predictions}}{\text{total predictions}}$ (\%).
We report the mean accuracy over 5 cross-validation folds, as mentioned earlier.

\subsubsection{Fetch Object}

In this task, the robot receives a natural language instruction to fetch an object, specifying its properties (e.g., {\it ``fetch an object that is cylindrical and short''}).
The robot is then presented with a group of objects, among which one matches the specified properties (i.e., target object), while the remaining distractor object(s) differ from the target object in at least one property.
To illustrate, if the robot is instructed to fetch an object that is both {\it cylindrical} and {\it short}, the distractor objects might be {\it cylindrical} or {\it short}, but not both.
The robot's objective is to interact with these presented objects and correctly identify one with the requested properties.
This task presents a challenge as the robot needs to detect the target object's properties given in natural language and distinguish it from the distractors by interaction. 
We evaluate the robot's performance on the fetch task across different levels of complexity.
In this task, we refer to the given instruction as a ``command'' and the objects presented to the robot are carefully chosen from the previously mentioned test set, ensuring that they are entirely new to the robot.
In difficulty \parai{Level 1}, the command specifies the category name of the target object (e.g., {\it ``fetch a ball''}); a distractor object is chosen from a different category.
In \parai{Level 2}, the command describes a specific property of the target object (e.g., {\it ``bring an object that is hard''}).
A distractor object is selected with a different property.
In the \parai{Level 3} scenario, the command includes two distinct properties of the target object (e.g., {\it ``bring an object that is small and hard''}).
The distractor object, on the other hand, possesses different properties.
For \parai{Level 4}, like Level 3, the command includes two target object properties.
However, this time, two distractor objects are introduced, each with differing properties.
\parai{Level 5} represents a variation of Level 2, where the commands only contain a property from a specific category, as illustrated in Table \ref{tab:property_categories}.
For instance, in the {\it ``Material''} category, the command might read, {\it ``get an object that is plastic.''}
Level 5 was introduced to assess the robot's performance across various property categories.
For each level, we created 20 commands for target objects and carefully selected corresponding distractor objects for each of the 5 previously explained folds.
For each object (target and distractor(s)), we calculated its selection percentage, defined as $S = \frac{\text{number of times the object is selected}}{\text{total number of commands}}$ (\%).
Our results are reported as the mean selection percentage across the 5 folds.

We employ the approach outlined in Algorithm \ref{alg:fetch_object} for this task.
Initially, we convert the natural language instruction into a text embedding, denoted as $t_c$, using CLIP's text encoder (step 1).
Subsequently, the robot interacts with the presented objects, including the target object and distractors, using various available behaviors while simultaneously recording sensory signals (step 5).
To simulate this step, we randomly select a trial from our dataset among 5 trials of each object.
Leveraging our trained framework, we generate unified representations, denoted as $u_b$, by processing the sensory inputs for each behavior (step 6).
Next, we calculate the cosine similarity between the command embedding ($t_c$) and the unified representation ($u_b$) for each behavior, maintaining a cumulative similarity score (step 7).
Finally, once all behaviors are considered, the object with the highest cumulative similarity score is identified as the target object, concluding the task (step 11).

\para{Baselines, Ablations, and Comparisons.} We ablate our full framework (MOSAIC), featuring the multi-sensory self-attention module, against a framework that omits this component (MOSAIC{\it-w/o-SA}).
We conduct these evaluations under two conditions: a non-interactive condition, where the robot solely performs the {\it Look} behavior, and an interactive condition, where the robot engages in all 9 aforementioned interactive behaviors.
\gt{Notably, in the {\it Look} behavior, only visual embeddings are employed as the unified representations after passing through the self-attention layer in MOSAIC, while, in MOSAIC{\it-w/o-SA}, the {\it Look} behavior utilizes only CLIP's vision encoder.}
Conversely, for interactive behaviors, all three modalities (i.e., visual, auditory, and haptic) are used to create unified representations.
For the object category recognition task, we report recognition accuracy separately for the {\it Look} behavior, each of the 9 interactive behaviors individually, and the combination of all 9 interactive behaviors.
The combined accuracy is calculated through a weighted combination of each behavior's performance on the training data.
We also compare our recognition accuracy with two baseline methods: Sinapov {\it et al.} \cite{sinapov_grounding_2014}, who trained a Support Vector Machine (SVM) classifier using handcrafted auditory, haptic features, and visual features, and Tatiya et al. \cite{tatiya2019deep}, who applied a deep learning approach to raw multi-sensory data for object category classification.
For the fetch object task (see Algorithm \ref{alg:fetch_object}), the set $B$ contains only the {\it Look} behavior for the non-interactive condition and all 9 interactive behaviors for the interactive condition.

\vspace{-0.2cm}
\section{Results}
\vspace{-0.1cm}
\begin{figure}
\centering
\includegraphics[width=0.95\linewidth]{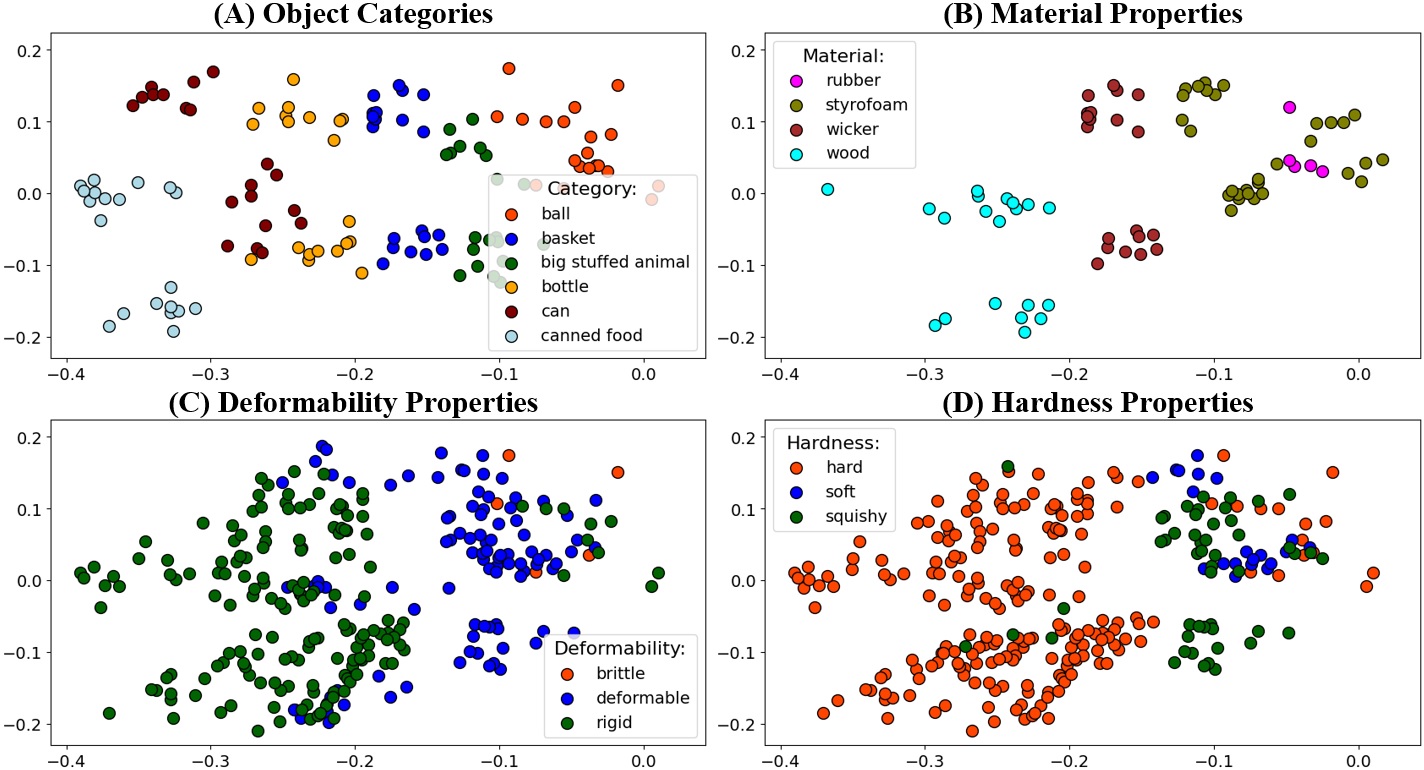}
\caption{\small 2D unified representations derived from autoencoder trained on {\it Push} behavior's data: (A) Object categories, (B) Material, (C) Deformability, and (D) Hardness properties.}
\label{fig:IE}
\vspace{-0.7cm}
\end{figure}

\para{An Illustrative Example.} Let's consider a scenario where the robot performs the {\it Push} behavior on 80 objects (4 objects x 20 categories), recording visual, acoustic, and haptic data.
With each object undergoing 5 trials, this yields a dataset of 400 examples (80 objects × 5 trials). Using our MOSAIC framework, we use this data to learn unified representations.
For visualization, we subjected these representations to dimensionality reduction using a linear autoencoder, resulting in a concise 2-dimensional latent space (Fig. \ref{fig:IE}).
This visualization encapsulates four object properties: object categories, material, deformability, and hardness.
Distinct colors are used to differentiate objects based on different values of these properties. To maintain clarity, we selectively plot only specific categories or objects with particular properties.
\gt{Due to the absence of certain properties in some objects, the observed inconsistencies in Fig. \ref{fig:IE} for different properties arise, as not every object possesses all the properties.}

These visualizations unveil meaningful insights.
Objects within the same category or material composition form tight clusters in the 2D space, showing the efficiency of our unified representations in capturing object semantics and material characteristics.
The deformability properties plot demonstrates a separation between {\it rigid} and {\it deformable} objects, with {\it brittle} ones inclining towards {\it deformable}.
Similarly, in the hardness properties plot, {\it hard} objects cluster on one side, while {\it soft} and {\it squishy} objects gravitate towards the opposite side.
Essentially, our unified representations effectively encode objects with similar properties, as evidenced by distinct clusters of similar objects, even when these objects belong to different categories or material groups across various property categories.
This illustrates MOSAIC's capacity to capture nuanced object attributes and relationships, a pivotal aspect of its performance across diverse tasks.

\para{Object Category Recognition Results.} Object category recognition results are presented in Table \ref{tab:category_recognition_accuracy_results}.
Note that the {\it Look} behavior only relies on visual modality, and the ``All behaviors'' row at the bottom refers to all 9 interactive behaviors combined.
Our approach, using unified representations, exhibits a remarkable level of competitiveness compared to state-of-the-art results for this dataset, demonstrating higher recognition accuracy in seven out of ten behaviors.
For the remaining three behaviors, we achieved comparable accuracy.
We achieved this level of performance using a straightforward linear model on top of the unified representations, a contrast to previous methods.
Notably, the prior work \cite{tatiya2019deep} employed a specialized neural network architecture tailored specifically for this task, while \cite{sinapov_grounding_2014} relied on handcrafted features.
Furthermore, our results consistently indicate that our full framework, including self-attention, outperforms the counterpart without self-attention.
This underscores the utility of the multi-sensory unified representation and the effectiveness of the self-attention mechanism in enhancing the robot's adaptability to diverse tasks.

\begin{table}
\caption{\small Category recognition accuracy (\%) 
for each behavior.}
\vspace{-0.1cm}
\label{tab:category_recognition_accuracy_results}
\footnotesize
\centering
\resizebox{0.95\columnwidth}{!}{
\begin{tabular}{@{}l@{\hspace{0pt}}c@{\hspace{5pt}}c@{\hspace{5pt}}c@{\hspace{5pt}}c@{\hspace{5pt}}}
{\it \textbf{Behavior}} 
    & {\it \textbf{Sinapov {\it et al.}\!\! \cite{sinapov_grounding_2014}}} 
    & {\it \textbf{Tatiya {\it et al.}\!\! \cite{tatiya2019deep}}} 
    & {\it \textbf{MOSAIC{\it-w/o-SA}}} 
    & {\it \textbf{MOSAIC (ours)}} \\
\toprule
Look & 67.7 & --- & 86.4 $\pm$ 1.2 & \textbf{87.4 $\pm$ 2.0} \\
\midrule
Grasp & 65.2 & 71.4 & 72.2 $\pm$ 6.7 & \textbf{74.0 $\pm$ 5.8 } \\
Hold & 67.0 & \textbf{76.8} & 68.0 $\pm$ 5.3 & 69.6 $\pm$ 5.2 \\
Lift & \textbf{79.0} & 77.8 & 72.8 $\pm$ 4.2 & 77.8 $\pm$ 5.7 \\
Drop & 71.0 & \textbf{78.0} & 73.2 $\pm$ 3.8 & 77.2 $\pm$ 5.9 \\
Poke & 85.4 & 73.8 & 81.6 $\pm$ 2.2 & \textbf{86.4 $\pm$ 1.0} \\
Push & 88.8 & 67.4 & 85.6 $\pm$ 3.5 & \textbf{89.4 $\pm$ 4.4} \\
Shake & 76.8 & 83.6 & 81.2 $\pm$ 6.2 & \textbf{84.0 $\pm$ 5.6} \\
Tap & 82.4 & 81.6 & 81.2 $\pm$ 5.7 & \textbf{84.4 $\pm$ 1.8} \\
Press & 77.4 & 58.8 & 71.6 $\pm$ 8.7 & \textbf{77.8 $\pm$ 6.4} \\
\midrule
All behaviors & --- & --- & 95.2 $\pm$ 3.6 & \textbf{95.6 $\pm$ 3.9} \\
\bottomrule
\end{tabular}
}
\vspace{-0.7cm}
\end{table}

\para{Fetch Object Results.} The fetch object task, whose results are summarized in Table \ref{tab:fetch_object_results}, comprises five distinct levels designed to assess the robot's ability to execute instructions.
In \parai{L-1} (Level-1), the command specified the object category name.
Our complete MOSAIC framework excelled in interactive behavior conditions, achieving an impressive target object selection rate of 99.0\%, outperforming all baseline models.
\parai{L-2 to L-5}: These levels introduced object properties into the command instead of specifying the object category name.
Generally, the interactive behaviors condition outperformed the non-interactive one, with our full MOSAIC model excelling in most cases.
Interestingly, providing more object properties in the command led to better performance, exemplified by a higher target object selection rate in L-3 compared to L-2, across all conditions, except for {\it ``Look''} without self-attention.
This suggests that learning unified representations with self-attention prioritizes the most relevant object properties.
\parai{L-4} presented greater challenges due to the inclusion of two distractor objects resembling the target object.
Nevertheless, our complete MOSAIC framework with self-attention consistently outperformed all baselines.

To evaluate the robot's ability to fetch objects based on specific property categories, we considered L-5, where the command included only descriptive words related to specific property categories.
For simplicity, we discuss five property categories.
\parai{Deformability and Weight}: In scenarios involving non-visual properties like deformability and weight, the interactive behaviors condition significantly outperformed the non-interactive one.
This aligns with intuition, as visual observation alone may not suffice to disambiguate these properties.
\parai{Transparency and Size}: For visual properties like transparency and size, the interactive behaviors condition performed comparably to the non-interactive condition, suggesting that interaction with objects may not yield significantly more information in these scenarios.
\parai{Shape}: Intriguingly, for the shape property category, the interactive behaviors condition significantly outperformed the non-interactive one.
This implies that interacting with objects enables the robot to observe them from various angles, enhancing its ability to predict object shape compared to merely observing from a top angle. In summary, our full MOSAIC framework demonstrated robust performance in the fetch object task, relying solely on unified representations without additional learning. These results underscore the adaptability and applicability of unified representations across diverse tasks, including those involving natural language instructions.

\begin{table}
\caption{\small MOSAIC's target object selection (\%) in various levels of the fetch object task, with and without Self-Attention.}
\vspace{-0.1cm}
\label{tab:fetch_object_results}
\footnotesize
\centering
\resizebox{0.9\columnwidth}{!}{
\begin{tabular}{l@{\hspace{2pt}}cccc}
 & \multicolumn{2}{c}{\it \textbf{Look (non-interactive)}} & \multicolumn{2}{c}{\it \textbf{Interactive}} \\
\cmidrule(r){2-3}
\cmidrule(r){4-5}
 & {\it-w/o-SA} & {\it \textbf{MOSAIC}} & {\it {\it-w/o-SA}} & {\it \textbf{MOSAIC }} \\
\toprule
{\textsc{Level 1}}  & 74 & 82 & 97 & \textbf{99} \\
{\textsc{Level 2}}  & 61 & 65 & \textbf{84} & 81 \\
{\textsc{Level 3}}  & 60 & 74 & \textbf{86} & 83 \\
{\textsc{Level 4}}  & 54 & 70 & 72 & \textbf{77} \\
\midrule
{\textsc{Level 5}} & & & & \\
\hspace{5pt}{\textsc{Deformation}} &  45 & 48 & 71 & \textbf{74} \\
\hspace{5pt}{\textsc{Shape}} &  85 & 80 & \textbf{97} & 95 \\
\hspace{5pt}{\textsc{Size}} & 62 & 74 & 72 & \textbf{75} \\
\hspace{5pt}{\textsc{Trasparency}} & 62 & 62 & 51 & \textbf{63} \\
\hspace{5pt}{\textsc{Weight}} & 52 & 63 & \textbf{85} & \textbf{85} \\
\bottomrule
\end{tabular}
}
\vspace{-0.7cm}
\end{table}

\section{Conclusion and Future Work}
\vspace{-0.1cm}
We introduced the MOSAIC framework to enable robots to generate versatile, multimodal representations through interactive object perception and to leverage these unified representations across various downstream robot learning tasks. 
Through extensive performance evaluation, we have showcased the effectiveness of these unified representations in tasks such as category recognition, using a simple linear probe setup, and the fetch object task under zero-shot conditions.
Moving forward, there are several exciting directions for future research. Firstly, we plan to consider the \textit{transfer} of unified representations across different robot morphologies, enabling a broader range of robots to benefit from this technology. Furthermore, we envision settings where interactive behaviors are learned and composed, alongside the tasks we considered in this paper, thereby further increasing the efficacy of object exploration. 
These future endeavors hold the potential to further enhance the utility of unified representations in robotics and expand their applications across a multitude of scenarios and environments.
One limitation in our current study is that, for the fetch object task, we evaluated using a zero-shot transfer condition rather than a learning-based approach to find the target object. For future work, it would be important to explore learning-based policies for solving the fetch object task, potentially increasing the versatility and adaptability of our framework.

\clearpage
\balance

\bibliographystyle{IEEEtran}
\bibliography{references}

\end{document}